\documentclass[a4paper,conference]{IEEEtran}
\usepackage{times}
\usepackage{epsfig}
\usepackage{graphicx}
\usepackage{amsmath}
\usepackage{amssymb}
\usepackage{xspace}
\usepackage{color}
\usepackage{array}
\usepackage{float}
\usepackage{textcomp}
\usepackage{multirow}
\usepackage{setspace}
\usepackage{subcaption}

\providecommand{\tabularnewline}{\\}
\floatstyle{ruled}
\newfloat{algorithm}{tbp}{loa}
\providecommand{\algorithmname}{Algorithm}
\floatname{algorithm}{\protect\algorithmname}

\ifCLASSINFOpdf
\else
\fi
\begin{document}
\title{Progressive Learning Algorithm for Efficient Person Re-Identification}


\author{\IEEEauthorblockN{Zhen Li}
\IEEEauthorblockA{Shanghai Grandhonor Information Technology  Co.Ltd\\
Nanjing University of Aeronautics and Astronautic\\
Shanghai 200072, China\\
Email: lizh0019@gmail.com}\\

\IEEEauthorblockN{Liang Niu}
\IEEEauthorblockA{New York University\\
New York University Abu Dhabi\\
New York, NY 10003\\
Email: liang.niu@nyu.edu}\\
\and
\IEEEauthorblockN{Hanyang Shao}
\IEEEauthorblockA{Shanghai Grandhonor Information Technology Co.Ltd\\
Shanghai 200072, China\\
Email: hansoluo757@gmail.com}\\
\\
\IEEEauthorblockN{Nian Xue*}
\IEEEauthorblockA{New York University\\
New York University Abu Dhabi\\
New York, NY 10003\\
Email: nian.xue@nyu.edu}\\
% \and
%\IEEEauthorblockN{Liangliang Cao}
%\IEEEauthorblockA{University of Massachusetts Amherst\\
%Amherst, MA, USA 01003\\
%Email: llcao@cs.umass.edu}
}

%Telephone: (800) 555--1212, Fax: (888) 555--1212}


\maketitle

\begin{abstract}
This paper studies the problem of Person Re-Identification (ReID)
for large-scale applications. Recent research efforts have been devoted
to building complicated part models, which introduce considerably
high computational cost and memory consumption, inhibiting its practicability
in large-scale applications. This paper aims to develop a novel learning
strategy to find efficient feature embeddings while maintaining the
balance of accuracy and model complexity. More specifically, we find
by enhancing the classical triplet loss together with cross-entropy
loss, our method can explore the hard examples and build a discriminant
feature embedding yet compact enough for large-scale applications.
Our method is carried out progressively using Bayesian optimization,
and we call it the Progressive Learning Algorithm (PLA). Extensive
experiments on three large-scale datasets show that our PLA is comparable
or better than the-state-of-the-arts. Especially, on the challenging
Market-1501 dataset, we achieve Rank-1=94.7\%/mAP=89.4\% while saving
at least 30\% parameters than strong part models.
\end{abstract}


% \fi
\IEEEpeerreviewmaketitle

\section{Introduction}
% structure of intro
% 1. analyze the problem: tradeoff of complexity and efficiency
% 2. insights: why triplet loss
% 3. our approaches. two major contributions: (branch after the pooling + bayesian batch hard examples)
% 4. our performance: highlight the performance

% llcao: reduce the length of trade off
% 1. insight of the problem: highlight the tradeoff 
One key challenge in person re-identification (ReID) is how to balance
the tradeoff between accuracy and model complexity. Person ReID aims
to retrieve a given person across a vast amount of videos, despite
significant variations under different surveillance cameras at different
locations. Some research studies \cite{Sun2017Beyond,Wang2018Learning}
have obtained the-state-of-the-art accuracy on public datasets, at
the cost of increasing computational complexity and a large number
of parameters. 
For example, MGN \cite{Wang2018Learning} splits the
network into one branch for global feature and two other branches
for local features, resulting in 180MB inference memory per image (or 23GB per batch of 128 images)
which prevents it from being used on resource-constrained platforms \cite{hsu2017fallcare+}.
%which is more than twice its ResNet-50 \cite{he2016deep} backbone,
%However, these advances to improve accuracy require much
%more computation and memory consumption, preventing it from being used in
%real-world systems. In many real-world applications such as large-scale
%video surveillance, the feature extraction tasks need to be carried
%out in a timely fashion on a resource-constrained platform, e.g. \cite{hsu2017fallcare+}.
% 4. For example, MGN  splits the network into one branch for global feature and two other branches for local features, resulting in roughly  multiplication  addition operations and 180MB inference memory per image that will consume at least 23GB GPU memory for a popular setting of batch size 128.
Another group of methods \cite{Sun2017SVDNet,Hermans2017In,Li2018Harmonious}
 focus on  scalable solutions that can support real-time search
across hundreds of hours of videos, with simpler network and compact
representation. However, the performance of the second group degenerates
quickly with the simpler model structure.

\begin{figure}[t]
\centering\includegraphics[scale=0.2]{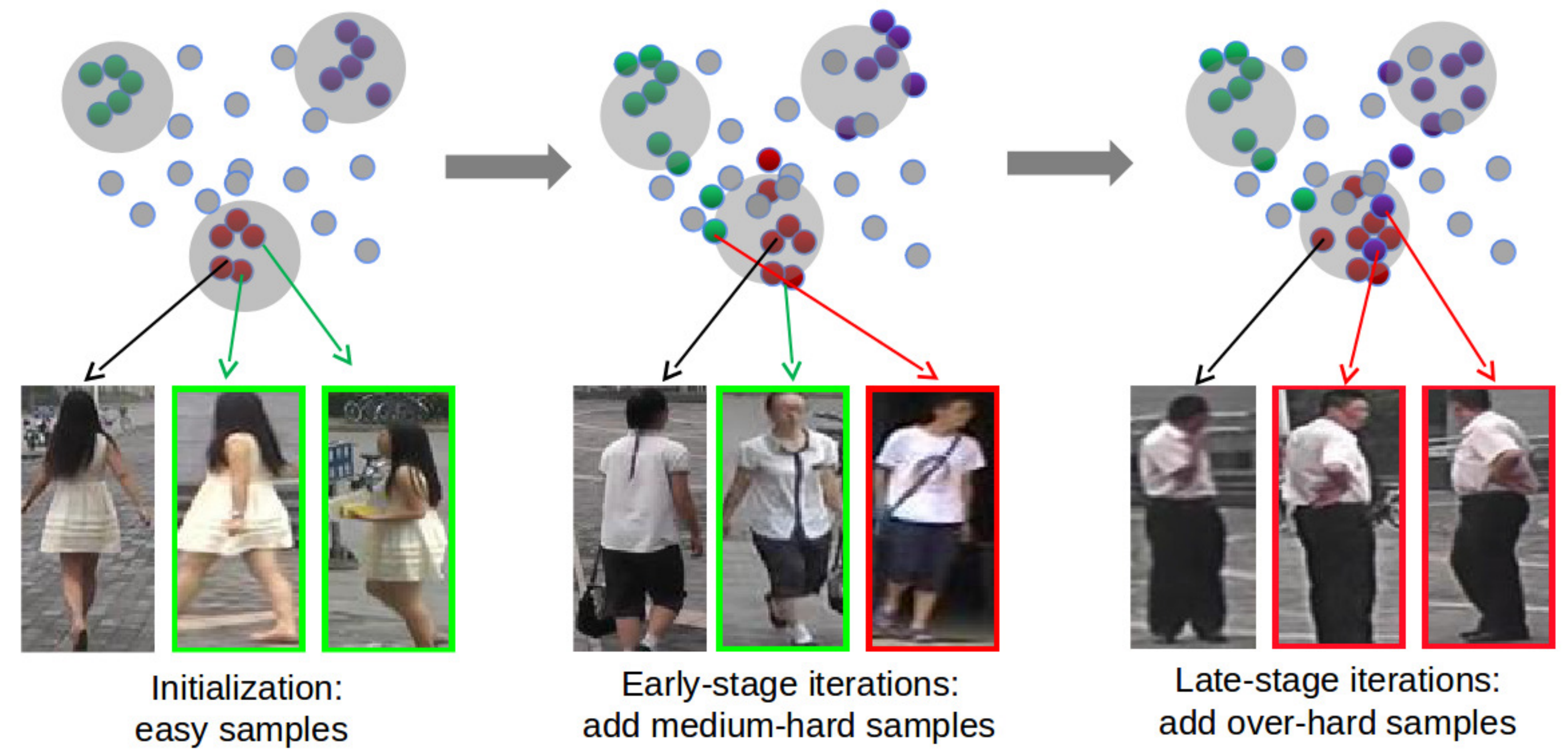} \caption{{ Training mini-batches consisting of easy, medium-hard and over-hard
examples. A red box indicates a person that is different from the
anchor sample, while a green box indicates the same person. The hard
examples are valuable to train a high precision model but may lead
to confusion in the optimization. We propose progressive learning
strategy to solve this challenge, which first learns easy and medium-hard examples. Note that training on over-hard examples is circumvented if they are less beneficial than medium-hard ones.} \label{fig:identity-pool} }
\end{figure}

% llcao:remove insight: coz reviewers are not impressed with end2end learning.
% 2 insight
%We believe it is possible to learn an efficient representation for
%person ReID that can match the-state-of-the-art performance. One very
%appealing strategy is to employ end-to-end-learning to find the optimal
%network. Classical deep learning methods usually learn the network
%as a multi-class classifier using cross entropy loss. Such a loss
%function expects all images of the same person will form a single
%cluster. This is not the case for person ReID, since one person may
%be captured from different viewpoints and represented by multiple
%body parts.

% 3 our approach
% 3.1 combine triplet loss and cross entropy loss
The goal of this paper is to learn an efficient model for
person ReID that can match the-state-of-the-art performance.
With the recent success  \cite{schroff2015facenet},
triplet loss is preferred by many researchers because it
does not require the images of the sample person to collapse to a
single cluster, so they can potentially model the variations and different
parts of the same person. However, triplet loss mainly focuses on the 
question of ``how similar are
the two images" \cite{Hermans2017In} instead of ``is this a new person", and often
suffers from a weaker generalization capability and slow convergence. 
In this paper, we believe triplet loss and  cross-entropy loss are  complementary to each other,
and propose to integrate two loss into one framework.

% 3.2 difficult in optimization
In practice, optimizing triplet loss at large scale is not
easy due to the difficulty of sampling good candidates of hard triplets.
When combining triplet loss with
cross-entropy loss, the optimization using stochastic gradient descents becomes more difficult. 
A straightforward
optimization may find an inferior solution or even fail to converge.
If we choose all the easy examples, the model may not be discriminant enough. 
But if we choose all hard examples, it will make it difficult for the Stochastic Gradient Descent (SGD) optimizer to get
out of the local minimum, and the model may stop improving in the early stage. 

% 3.3 PLA
We believe the key to solve the difficult optimization problem is to train the model in a progressive way.
At the beginning of the stochastic optimization process, we get more simple samples  
in every batch, to make sure that most of them could be recognized correctly. In the later,
we focus on the hard examples, and reduce the number of simple samples. 
The resulted algorithm is to progressively
learn the triplet loss from simple samples to hard samples using the Bayesian framework. 
We call this new approach as Progressive Learning Algorithm (PLA).
Fig. \ref{fig:identity-pool} shows the progressive learning in different strategies. 
This approach can help to optimize not only triplet loss but also the new composite loss for ReID problems. 

%To solve the difficulty of optimizing the new loss function, 
% this paper develops a novel Bayesian approach, which is employed to progressively learn 
% the triplet loss from simple samples to hard samples
%and is combined with the traditional cross entropy loss. 

%illustrates identities of three levels
%of difficulty. It can be noticed that training only the hard identity
% examples will confuse the CNN weights and might result in failure
%in associating the easy identity examples to the triplet anchor, while
%training the medium-hard examples will probably result in better ReID
%performance.

%Note that the Bayesian-based progressive learning technique is orthogonal
%with other architectural novelties so we expect the cost function
%and efficient network structure in this paper can be used together
%with other methods in different scenarios.

% 3.4 integrate different branches
Using the framework of PLA, this paper found an efficient way to integrate
different branches in the ReID network structure. We used one shared
convolutional network (i.e., the ResNet-50 \cite{he2016deep}) for
each image. Note that almost all previous works used different convolutional
networks for separate branches. The proposed architecture significantly
reduces the computational cost and model complexity. The backbone
is not altered and the feature maps after global pooling are fed to
$1\times1$ convolution, producing two global features which are evaluated
using cross-entropy loss and triplet loss, respectively. It is convenient
to deploy the inference model due to its compact nature.

% wrong terms: no softmax loss. unclear batch hard loss = PLA?
%The classification loss is implemented using Softmax loss, while the triplet loss is realized by the proposed generalized batch hard loss.

% 3.5 performance
Our approach enjoys a good balance of
model complexity and accuracy. Our model is almost as efficient as
ResNet-50 based DaRe \cite{wang2018resource}, and can finish inference
(batch size 128) of 17,000 images in a second, using 1 GTX 1080Ti card,
while still obtaining comparable accuracy with the-state-of-the-arts. On the challenge Market1501 dataset, our single query model
without reranking achieves the Rank-1 score of $93.7\%$. %and retrieval of at least 16,000 images
%\llcao{Is it fast enough? I feel it cannot scale if you have hundreds of hours of videos}
%\llcao{please add more details : how many instances are detected per image/frame,  how fast is the decoding, and how fast is the retrieval}

% optional:
This paper is structured as follows. In Section 2, we review the related
works. The proposed PLA is elaborated in Section 3, followed by the
extensive experiments in Section 4 which validates the superiority
of PLA in the balance of accuracy and resource requirement. Finally,
Section 5 concludes this work.

\begin{figure}[tb]
\centering \includegraphics[scale=0.22]{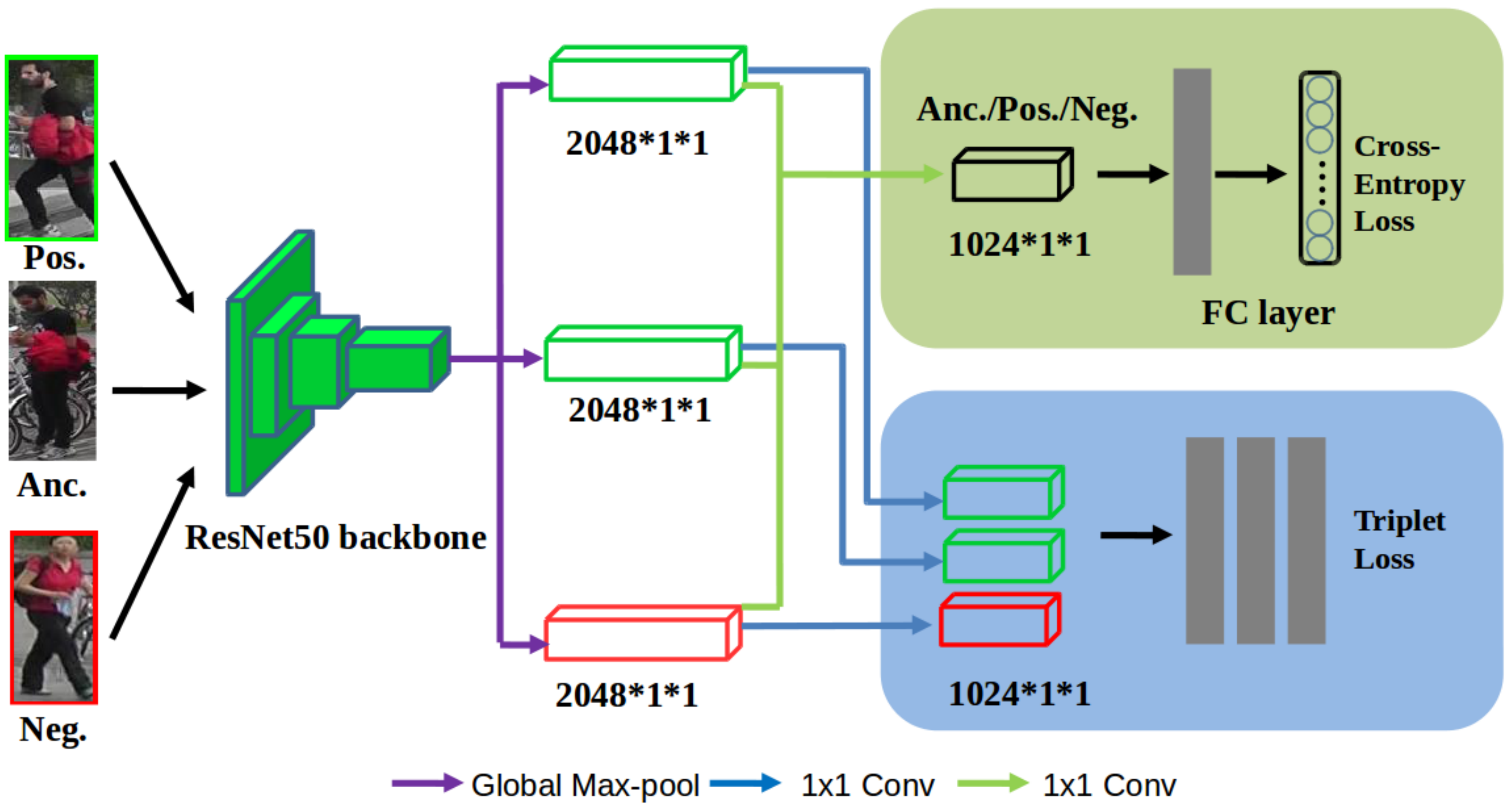} \caption{Overview of the proposed ReID network architecture. ``Anc.'', ``Pos.''
and ``Neg.'' represent anchor image, positive images that belong
to the same identity and negative images that belong to different
identities, respectively. \label{fig:framework} }
\end{figure}

\section{Related Work}

Deep learning models for ReID can be categorized into global models
and part models. Global models utilize only global features, typically
after global pooling operations, without analyzing regions of feature
maps. In contrast, part models typically learn various local features
in spatial domain of feature maps.

\subsection{Global model}

A breakthrough for the ReID task is triplet loss \cite{weinberger2006distance} by which the CNN
network weights are optimized until the features belonging to each
identity lie closer compared to features from different identities.
The main difficulty of learning person ReID models using the triplet
loss is the mining of hard triplets, as otherwise it will frequently
produce disappointing results, or even worse, selecting over-hard
triplets will result in an unstable training process \cite{Hermans2017In}.

So far, a large number of variants of the triplet loss have been proposed
to improve the ReID performance \cite{oh2016deep,khamis2014joint,ding2015deep,paisitkriangkrai2015learning,Cheng2016Person,Wang2016Joint,shi2016embedding,Su2016Deep,chen2017multi,Liu2016End,Hermans2017In}
regardless of the network structure. The triplet loss has become the
$de\ facto$ standard of the loss function in ReID deep learning networks.
In \cite{wang2018resource}, Deep Anytime Re-ID (DaRe) is proposed
which is a resource-aware ReID architecture that combines pooled feature
maps from multiple convolutional network layers. The composite loss
function consists of per-stage losses and final-stage triplet loss.
The ResNet-50 based DaRe approach achieves $88.5\%$ Rank-1 accuracy
on the Market-1501 dataset, which is a milestone of the ReID approaches
based on global feature. \textcolor{black}{GP-reid \cite{almazan2018re}
discusses a set of good practices for person ReID, especially for
using global features and triplet loss}, and achieves promising results.
In this work, we will show that, by generalizing the triplet loss,
it has more potential for learning ReID model by global features.

\subsection{Part model}

In order to relieve the problem of detection misalignment and background
clutter, a rich body of works have tried to utilize local features
in deep learning networks, with the entire CNN parameters being optimized
in an end-to-end manner \cite{li2017learning,Zhao2017Deeply,Su2017Pose,Cheng2016Person,li2017person,Sun2017Beyond,Zhao2017Spindle,Yao2017Deep,Liu2017HydraPlus,Li2018Harmonious}.

The main idea of attention model is to evaluate region-level or pixel-level
attention map, and implement the importance weighing on feature maps
in the middle of the CNN network before global pooling. For region-level
attention \cite{li2017learning,Zhao2017Deeply,Su2017Pose}, some works
utilize empirical knowledge about human bodies \cite{Cheng2016Person,li2017person,Sun2017Beyond,zhang2017alignedreid},
some acquire region by region proposal methods \cite{Yao2017Deep,li2017learning}
while others do not rely on the structural information or region proposal
but the attention task is achieved during deep learning \cite{Su2017Pose,Zhao2017Spindle}.
In order to further reduce the impact of the background clutter, some
attention models are proposed to locate the body part in either middle-level
or fine-grained pixel-level \cite{Zhao2017Deeply,Liu2017HydraPlus,Liu2016End,Li2018Harmonious}.
The attention models typically improve ReID performance but require
more computation and memory. 

As an alternative to implement part models, a multi-branch architecture
has been proposed to learn an ensemble of network branches which improves
the accuracy compared with any single branch setting \cite{zhang2017alignedreid,Sun2017Beyond,Wang2018Learning,Li2018Harmonious,Bai2017Deep}.
A typical work is Part-based Convolutional Baseline (PCB) \cite{Sun2017Beyond}
which applies uniform partition strategy and concatenates region-level
local features to form the output composite feature.  The
authors of \cite{Li2018Harmonious} proposed a multi-branch architecture
HA-CNN in which the global branch learns the global feature while
each of the three local branches aims to learn the discriminative
local features for its corresponding image regions.  Deep-Person \cite{Bai2017Deep} is proposed
to model the spatial dependency of body parts by Long Short-Term Memory
(LSTM) in an end-to-end way. The recent work Multiple Granularity
Network (MGN) \cite{Wang2018Learning} consists of one global and
two local branches from $res\_conv4$ layer with independent parameters,
which can boost each other. MGN can achieve outstanding performance
exceeding all previous methods, achieving a milestone of $96.6\%$
Rank-1 accuracy on Market-1501 dataset with a reranking technique.
However, the multi-branch architecture usually requires even more
computation and memory resources than the part-model based attention
methods impeding its practicality in deployment.

\section{Learning Composite Feature Embedding}

\subsection{Network Architecture}

In the proposed plain CNN architecture, the ResNet-50 \cite{he2016deep}
backbone is used for learning the main discriminative ability. 
The architecture
is shown in Fig. \ref{fig:framework}, where the feature vectors after
global pooling are fed to $1\times1$ convolution, producing 1024-dim
features. These features are evaluated using cross-entropy loss and
triplet loss, respectively, and concatenated to form the final 2048-dim
feature embeddings. The details of loss function and optimization will
be discussed in subsection \ref{subsec:generalized-triplet-loss} and \ref{subsec:Bayesian-Optimization}.

We have conducted preliminary experiments on a single embedding, i.e.,
the 2048-dim feature after global pooling, with both losses. However,
the performance is not as good as training two separate embeddings
with the two different loss functions respectively. The main reason
is that cross-entropy loss and triplet loss are formulated quite differently, and two heads after global pooling are necessary to boost each other.

\subsection{ Loss Function for Progressive Learning\label{subsec:generalized-triplet-loss}}

As commonly defined in the triplet loss based feature embedding methods
\cite{schroff2015facenet,shi2016embedding,Hermans2017In}, the goal
of metric embedding learning is to learn a mapping from image data
to semantic features.
%$g\left(\boldsymbol{\theta}\right):\mathbb{R}^{D}\rightarrow\mathbb{R}^{F}$
%, where $\mathbb{R}^{D}$ denotes the image data manifold space, $\mathbb{R}^{F}$
%denotes the semantic feature space, $\boldsymbol{x}$ denotes the input data.
We use $\boldsymbol{\theta}$ to denote
the network parameters.
The function $g\left(\boldsymbol{\theta}\right)$ is usually implemented
using CNN, which is a non-linear mapping. The metric measure is defined
as 
$D_{i,j}=D\left(g_{\boldsymbol{\theta}}\left(\boldsymbol{u}\right),g_{\boldsymbol{\theta}}\left(\boldsymbol{v}\right)\right):\mathbb{R}^{D}\times\mathbb{R}^{D}\rightarrow R$. 

The original form of triplet loss is defined as a large margin nearest
neighbor (LMNN) loss \cite{weinberger2006distance}:
\begin{align}
\mathcal{L}_{LMNN}\left(\boldsymbol{\theta};X\right) & =\left(1-\mu\right)\sum_{y_{a}=y_{b}}D_{a,b}\label{eq:LMNN}\\
 & +\mu\sum_{\underset{y_{a}=y_{b}\neq y_{n}}{a,b,n}}\left[m+D_{a,b}-D_{a,n}\right]\nonumber 
\end{align}
where $a$ and $b$ denote samples in the same class, and $\mu$ and $m$ denote the weighting factor and the margin
parameter, respectively. 
The LMNN loss is essentially a weighted sum of
the intra-class Euclidean loss and a variation of the hinge loss. 
%commonly used in SVM \cite{rosasco2004loss}. 
%Ideally, by using such loss function,
%$g\left(\boldsymbol{\theta}\right)$ should map semantically similar
%images onto metrically close feature points, and semantically different
%points onto metrically distant points.

%A number of variants were proposed \cite{shi2016embedding,oh2016deep,Hermans2017In}
%following the LMNN loss, and amongst them, the ``batch hard'' triplet
%loss proposed in \cite{Hermans2017In} outperforms all previous triplet
%loss variants. 
Hermans et al. \cite{Hermans2017In} extended triplet loss 
by introducing the ``batch hard" triplet loss which is defined as follows:
\begin{equation}
\mathcal{L}_{BH}\left(\boldsymbol{\theta};X\right)=\sum_{l=1}^{P}\sum_{\underset{y_{a}=y_{b}=l}{a,b}}^ {}\left[m+\mathop{{\rm max}}D_{a,b}-\mathop{{\rm min}}\limits _{y_{n}\neq y_{a}}D_{a,n}\right]_{+}\label{eq:BH}
\end{equation}
where $a$ and $b$ belong to the same label $y_{a}\in\left[1,\dots,P\right]$,
$n$ belongs to a different label with $y_n\neq y_a$. Each label contains $K$
images, forming a batch of $P\times K$ images for training. Note that the batch parameters
$K$ and $P$ are constrained by the memory size, which may limit
the discriminative ability of the triplet loss strategy. The selected
triplets are derived from a small subset of the input data, thus the
``hardest'' (defined by $max$ and $min$ operations for hardest
positive and hardest negative, respectively) can be considered as
moderate triplets, which is neither too hard nor too easy for learning
with the triplet loss.

%However, this is still half on the way for the purpose of defining
%what is a good triplet. One obvious reason is that, the hardest examples
%selected in a subset of the input data may also be the hardest out
%of the whole input data, rendering the training process unstable \cite{Hermans2017In}.

In this work, we introduce two \textbf{progressive parameters} $p$ and $k$ to control the 
optimization of triplet loss.  Suppose $k\in\left[1,\dots,K\right]$, $p\in\left[1,\dots,P\right]$,
and $\mathop{{\rm Largest}^{k}\left(\cdot\right)}$ and $\mathop{{\rm Smallest}^{p}\left(\cdot\right)}$
denote the operations to find the $k$-th largest value of $D_{a,b}$ in the batch and $p$-th
smallest values of $D_{a,n}$, respectively. 
%\llcao{When and how often are these values evaluated? do we calculate such record on the fly or keep it in GPU memory?} 
Then we can generalize the batch hard triplet loss as
\begin{equation}
\mathcal{L}_{GBH}^{k,p}\left(\boldsymbol{\theta};X\right)=\sum_{l=1}^{P}\sum_{\underset{y_{a}=y_{b}=l}{a,b}}^ {}\mathop{{\rm ln}}\left(1+e^{m+T_{k,p}\left(a,b,n\right)}\right)\label{eq:GBH}
\end{equation}
\begin{equation}
T_{k,p}\left(a,b,n\right)=\mathop{\mathop{{\rm Largest}^{k}}}\left(D_{a,b}\right)-\mathop{{\rm Smallest^{p}}}\limits _{y_{n}\neq y_{a}}\left(D_{a,n}\right)\label{eq:generalized-distance}
\end{equation}
 The generalized batch hard loss can
define triplets at any hard level, by adjusting the parameters $k$
and $p$. When $k=p=1$ the generalized batch hard loss is exactly
the same as the batch hard loss defined in Eq. \ref{eq:BH}. When
$k>1$,  easier examples are selected for training, thus
the unstable training situation is alleviated.

%\subsection{Composite Loss \label{subsec:Loss}}

%For the lack of direct supervision from ground truth labels, simply
%using a triplet loss might sometimes result in non-optimum compared
%to the cross-entropy loss. As suggested
%in \cite{chen2017multi,zheng2016person,li2017person}, a multi-loss
%function is desired to update the CNN weights. We further integrate
%the cross-entropy loss to smooth/improve the solution and alleviate
%effects of the too hard examples. The cross-entropy loss is implemented
%with Softmax function which is defined as:

%\begin{equation}
%\mathcal{L}_{softmax}\left(\boldsymbol{\theta};X\right)=-\sum_{i=1}^{PK}\mathop{{\rm %ln}\frac{e^{W_{y_{i}}^{T}\boldsymbol{f}_{i}+b_{y_{i}}}}{\sum_{j=1}^{M}e^{W_{y_{j}}^{T}\boldsymbol{f}_{j}+b_{y_{j}}}}}\label{eq:softmax}
%\end{equation}
%where $M$ is the number of identities in the training dataset, and
%$W$ denotes the $1\times1$ convolution weights for Softmax function.
    
Our final objective is  the composite loss with the cross-entropy
loss and the generalized batch hard triplet loss as follows:
\begin{equation}
\mathcal{L}^{k,p}\left(\boldsymbol{\theta};X\right)=\mathcal{L}_{softmax}\left(\boldsymbol{\theta};X\right)+\lambda\mathcal{L}_{GBH}^{k,p}\left(\boldsymbol{\theta};X\right)\label{eq:Total-Loss}
\end{equation}
where $\lambda$ is a predefined weight for generalized triplet
loss as an regularization to the conventional cross-entropy loss.
In order to determine its value, we have made a joint Bayesian optimization
on parameters in subsection \ref{subsec:Bayesian-Optimization}, and
the $\lambda\in\left[0,2\right]$ is set dynamically.

In practice,  our loss function can be calculated efficiently. Following \cite{Hermans2017In},
we form batches by randomly sampling $P$ classes and randomly select $K$ images of each class.
Given $PK$ embeddings, we can compute their pairwise distance and sort them in $O(PK \log (PK))$. Then for each sample, we can select the $k$-th largest value from all positive distances $D_{a,b}$ with $y_a=y_b$, and $p$-th smallest value from all negative distances $D_{a,n}$
with $y_a\neq y_n$. In our experiments $PK$ is as small as 128, for which the sorting algorithm finishes in a short time. Overall our computation cost is comparable with that in \cite{Hermans2017In}.

\subsection{Progressive Learning Based on Bayesian Optimization \label{subsec:Bayesian-Optimization}}

As we can see from the previous section, the progressive parameters $p$ and $k$ are crucial to train the ReID model. When $k=8$, the optimizer selects
easy examples for every batch. When $k=1$, the hardest example is selected for optimization. 

Intuitively we could choose $k=8$ at the beginning of the optimization and then $k=1$  later for progressive learning. This incremental nature allows the training to first discover a stable solution for easy triplets and then shift attention to increasingly finetune weights to discriminate hard triplets, while avoiding over-hard triplets if they have no benefits.  However, such a holistic approach cannot  decide when and how often we shall adjust $k$, and cannot suggest whether we shall adjust other hyper-parameters $\lambda, m$ accordingly. In this paper, we employ Bayesian optimization \cite{shahriari2016taking} to adjust $p,k$ as well as the other hyper-parameters. Bayesian optimization provides a general framework for minimization of non-convex objective functions,
by which the parameters can be jointly optimized by iteratively updating the probabilistic model based on previous
exploration.

%Our primary contribution is a training methodology for ReID where we start with easier triplets, and then progressively increase the hardness by controlling the hyper-parameters of the proposed loss. This incremental nature allows the training to first discover a stable solution for medium-hard triplets and then shift attention to increasingly finetune weights to discriminate hard triplets, while avoiding over-hard triplets if they have no benefits. Progressive learning has showed its success in other areas, e.g. \cite{karras2017progressive}.
Suppose $\boldsymbol{w}=(\lambda, m, k, p)$ denotes all the hyper-parameters for our progressive learning. 
Define $f\left(\boldsymbol{\theta};\boldsymbol{w};X\right)$ as the
non-linear function that maps CNN parameters $\boldsymbol{\theta}$,
hyper-parameters $\boldsymbol{w}$, and input data $X$ to the objective
function. As Bayesian optimization is used here to determine the hyper-parameters
only, $f\left(\boldsymbol{\theta};\boldsymbol{w};X\right)$ is briefed
as $f\left(\boldsymbol{w}\right)$ in this work. Assume $f\left(\boldsymbol{w}\right)$
follows a Gaussian process which is parameterized by a mean function
$\mu\left(\cdot\right)$ and a covariance kernel ${\rm K}\left(\boldsymbol{w}_{1},\boldsymbol{w}_{2}\right)$
%\llcao{confusing: K is used both as a kernel function or 8 in the batch learning?}
:
$f\sim\mathcal{GP}\left(\mu\left(\cdot\right),{\rm K}\left(\cdot,\cdot\right)\right)$,
then its mean and covariance are $\mu\left(\boldsymbol{w}\right)=\mathbb{E}\left[f\left(\boldsymbol{w}\right)\right]$
and ${\rm K}\left(\boldsymbol{w}_{1},\boldsymbol{w}_{2}\right)=\mathbb{E}\left[\left(f\left(\boldsymbol{w}_{1}\right)-\mu\left(\boldsymbol{w}_{1}\right)\right)\left(f\left(\boldsymbol{\boldsymbol{w}}_{2}\right)-\mu\left(\boldsymbol{\boldsymbol{w}}_{2}\right)\right)\right]$,
respectively. By Gaussian distribution, ${\rm K}_{{\bf B}}\left({\bf \boldsymbol{w}}_{1},\boldsymbol{w}_{2}\right)=\frac{1}{\left(2\pi\right)^{\frac{d}{2}}\left|{\bf B}\right|^{\frac{1}{2}}}exp\left(-\frac{1}{2}{\bf \boldsymbol{\boldsymbol{w}}}_{1}^{T}{\bf B}^{-1}{\bf \boldsymbol{\boldsymbol{w}}}_{2}\right)$,
and ${\bf B}$ is the bandwidth matrix calculated from given samples.

Denote ${\bf \boldsymbol{\boldsymbol{w}}}_{i}=\left(\lambda_{i},m_{i},k_{i},p_{i}\right)$
as a set of hyper-parameters.
%\llcao{I changed this sentence. previously ``for each mini-batch $i$ in the network". please check whether it is correct.}.
For $N$ sets of such parameters we denote $\mathbb{W}=\left\{ \boldsymbol{w}_{1},{\bf \boldsymbol{w}}_{2},\cdots,\boldsymbol{w}_{N}\right\}$ ,
and the corresponding  $f\left(\mathbb{W}\right)=\left\{ f\left(\boldsymbol{w}_{1}\right),f\left({\bf \boldsymbol{w}}_{2}\right),\cdots,f\left(\boldsymbol{w}_{N}\right)\right\} $.
The posterior belief of $f$ at a new candidate ${\bf \hat{\boldsymbol{w}}}$
is given by
\begin{equation}
\begin{cases}
\begin{array}{c}
\tilde{f}\left({\bf \hat{\boldsymbol{w}}}\right)\sim\mathcal{GP}\left(\mu\left(\hat{\boldsymbol{\boldsymbol{w}}}\right)+\triangle\mu,{\rm \mathcal{K}}\left({\bf \hat{\boldsymbol{\boldsymbol{w}}}}\right)-\triangle{\rm \mathcal{K}}\right)\\
\triangle\mu={\rm \mathcal{K}}\left({\bf \hat{\boldsymbol{w}}},\mathbb{W}\right){\rm \mathcal{K}}\left(\mathbb{W}\right)^{-1}\left(f\left(\mathbb{W}\right)-\mu\left(\mathbb{W}\right)\right)\\
\triangle{\rm \mathcal{K}}={\rm \mathcal{K}}\left({\bf \hat{\boldsymbol{\boldsymbol{w}}}},\mathbb{W}\right){\rm \mathcal{K}}\left(\mathbb{W}\right)^{-1}{\rm \mathcal{K}}\left(\mathbb{W},{\bf \hat{\boldsymbol{w}}}\right)
\end{array}\end{cases}\label{eq:posterior}
\end{equation}
where ${\rm \mathcal{K}}\left(\mathbb{W}\right)={\rm \mathcal{K}}\left(\mathbb{W},\mathbb{W}\right)$
for brief. 

%\llcao{this algorithm's layout looks like main text. better to move to main text?}
\begin{algorithm}[tb]
\noindent \textcolor{black}{\caption{Training the ReID model based on PLA. \label{PL-REID} }
}

\begin{singlespace}

\textbf{\textcolor{black}{Input}}: A fixed-size mini-batch consisting of $P=16$ randomly selected identities and $K=8$ randomly selected images per
identity from the training set. 

\textbf{\textcolor{black}{Output}}: The optimal hyperparameter ${\bf \boldsymbol{\boldsymbol{w}}}^{*}$
along with the well trained CNN. %Final embedded feature for the $i$-th image for the $j$-th identity.

\textbf{\textcolor{black}{Initialization}}: Randomly initialize $N$ sets
of hyper-parameters $\mathbb{W}=\left\{ \boldsymbol{w}_{1},{\bf \boldsymbol{w}}_{2},\cdots,\boldsymbol{w}_{N}\right\} $ 
where ${\bf \boldsymbol{\boldsymbol{w}}}_{i}=\left(\lambda_{i},m_{i},k_{i},p_{i}\right)$,
$\lambda_{i}\in\left[0,2\right]$, $m_{i}\in\left[-0.1,0.3\right]$,
$k_{i}\in\left[1,8\right]$, $p_{i}\in\left[1,16\right]$ for $i=1,\cdots,N$. 

\end{singlespace} 

\noindent \textbf{Repeat}

\noindent \,\,\,\,\textbf{for} each hyperparameter $i=1$ to $N$
\textbf{do}

\noindent \,\,\,\,\,\,\,\,Exploration: Backpropagate CNN in
20 epochs and evaluate the loss $\mathcal{L}$ according to Eq. \ref{eq:GBH}
and Eq. \ref{eq:Total-Loss}, and evaluate the Bayesian objective
$f\left(\boldsymbol{w}_{i}\right)$. % according to Eq. \ref{eq:Objective};

\noindent \,\,\,\,\,\,\,\,Restoration: CNN weights are restored
to that before 20 epochs of exploration.

\noindent \,\,\,\,\textbf{end for}

\noindent \,\,\,\,Exploitation: Based on $f\left(\mathbb{W}\right)$
, obtain a new improved candidate ${\bf \boldsymbol{w}^{'}}$ and
update Gaussian process according to Eq. \ref{eq:posterior} and Eq.
\ref{eq:closedform}, and add ${\bf \boldsymbol{w}^{'}}$ to $\mathbb{W}$. 
%and update $\hat{\boldsymbol{w}}$ as well; 
%\llcao{Add ${\bf \boldsymbol{w}^{'}}$ to $\mathbb{W}$. }

\noindent \,\,\,\,Backpropagate to update CNN weights for 300
epochs based on the new hyperparameter ${\bf \hat{\boldsymbol{w}}}$
and the feed-forward loss $\mathcal{L}$;

\noindent \,\,\,\,Save the model with lowest loss $\mathcal{L}$ for the current hyperparameter ${\bf \hat{\boldsymbol{w}}}$; 

\noindent \textbf{Until} maximum epochs ($M=3,000$) reached 
\end{algorithm}
Let ${\bf \boldsymbol{\boldsymbol{w}}^{'}}$ denote the best candidate
evaluated so far, then the expected improvement of a candidate $\hat{\boldsymbol{\boldsymbol{w}}}$
is defined as
\begin{equation}
\mathcal{EI}\left({\bf \hat{\boldsymbol{\boldsymbol{w}}}}\right)=\mathbb{E}\left[f\left({\bf \boldsymbol{\boldsymbol{w}}}^{'}\right)-\tilde{f}\left({\bf \hat{\boldsymbol{\boldsymbol{w}}}}\right)\right]_{+}
\end{equation}
which can be efficiently computed in closed form:
\begin{equation}
\mathcal{EI}\left({\bf \hat{\boldsymbol{\boldsymbol{w}}}}\right)=\left({\rm \mathcal{K}}\left({\bf \hat{\boldsymbol{\boldsymbol{w}}}}\right)-\triangle{\rm \mathcal{K}}\right)^{\frac{1}{2}}\left(Z\varPhi\left(Z\right)+\phi\left(Z\right)\right)\label{eq:closedform}
\end{equation}
where $Z=\left(\tilde{\mu}_{f}\left({\bf \hat{\boldsymbol{\boldsymbol{w}}}}\right)-f\left({\bf \boldsymbol{\boldsymbol{w}}}^{'}\right)\right)/\left({\rm \mathcal{K}}\left({\bf \hat{\boldsymbol{\boldsymbol{w}}}}\right)-\triangle{\rm \mathcal{K}}\right)^{\frac{1}{2}}$,
and $\varPhi\left(\cdot\right)$ and $\phi\left(\cdot\right)$ are
the standard normal cumulative distribution function and probability
density function, respectively. 

In each iteration of Bayesian optimization, a fixed-size of new hyper-parameters are sampled where the one that maximizes $\mathcal{EI}\left({\bf \hat{\boldsymbol{w}}}\right)$ is chosen, and $\mathbb{W}$ is updated accordingly. For more details, one can refer to \cite{snoek2012practical}. 

%\llcao{I don't understand where $\mathcal{EI}$ is ever used. Update W?}

%\subsection{Objective Function}\llcao{I'd merge this subsection with above}

%using Bayesian optimization, where $\lambda_{i},m_{i},k_{i},p_{i}$ are regularization factor in
%Eq. \ref{eq:Total-Loss}, margin value in Eq. \ref{eq:GBH}, and indices
%of the hard positive and negative examples in Eq. \ref{eq:GBH}.  

Then the objective of the Bayesian optimization is determined by loss function in the last
section. In practice, we find when the composite loss decreases in a speed of relatively 15\% 
of the previous loss, the training is  healthy with good hyper-parameters. So in our system,
Bayesian optimization minimizes the following objective function: 
\begin{equation}
f\left(\boldsymbol{\boldsymbol{w}}\right)=\lvert\frac{\bar{\mathcal{L}}_{t}^{k,p}\left(\boldsymbol{\theta};\boldsymbol{\boldsymbol{w}};X\right)-\bar{\mathcal{L}}_{t^{'}}^{k,p}\left(\boldsymbol{\theta};\boldsymbol{\boldsymbol{w}};X\right)}{\bar{\mathcal{L}}_{t}^{k,p}\left(\boldsymbol{\theta};\boldsymbol{\boldsymbol{w}};X\right)}-\mathcal{ED}\rvert\label{eq:Objective}
\end{equation}
%\llcao{is the $f$ the same as the one in eq (6)?}
where $\bar{\mathcal{L}}_{t}^{k,p}\left(\boldsymbol{\theta};\boldsymbol{\boldsymbol{w}};X\right)$ and $\bar{\mathcal{L}}_{t^{'}}^{k,p}\left(\boldsymbol{\theta};\boldsymbol{\boldsymbol{w}};X\right)$
are the average CNN loss functions in the first 10 epochs and second 10 epochs of the $t$-th exploration in Algorithm \ref{PL-REID},  $\mathcal{ED}$
is the expected drop rate of the loss, which is empirically set to
0.15 as a healthy training process. 
%The optimal hyper-parameters are
%finally determined as ${\bf \boldsymbol{\boldsymbol{w}}^{*}}=\mathop{{\rm arg}\;{\rm min}}\limits %_{\boldsymbol{\boldsymbol{w}}}\left(f\left(\boldsymbol{\boldsymbol{\boldsymbol{w}}}\right)\right)$.

The overall PLA is shown in Algorithm \ref{PL-REID}.
The whole training procedure is end-to-end with the backbone of ResNet-50 which
is pretrained from ImageNet, and the $1\times1$ convolutions after
global pooling are randomly initialized. Based on the trained PLA
model, a 2048-dim feature will be generated from a query image to
match features extracted from the gallery images.

\section{Experiments}

In this section, we first introduce datasets and evaluation protocols
used in our experiments, then parameter setting and performance analysis
are provided. This is followed by experimental results compared with
other state-of-the-art methods. %Since pre-trained models often obtain great scores for person ReID \cite{Geng2016Deep,Zheng2016A}, while ever fewer top-performing approaches use networks trained from scratch \cite{li2014deepreid,Ahmed2015An,Cheng2016Person,Xiao2016Learning,shi2016embedding,Varior2016Gated}, we use the pre-trained model of ResNet-50 \cite{he2016deep}. 
A reranking technique \cite{Zhong2017Re} is a popular option to post-processing to improve the ReID precision.

\subsection{Datasets and Protocols}

We perform all the experiments on three commonly used benchmarks:
Market-1501 \cite{zheng2015scalable}, DukeMTMC-ReID \cite{ristani2016MTMC}
(briefed as DukeMTMC), and CUHK03(D) \& CUHK03(L) \cite{li2014deepreid} datasets.
These ReID datasets are summarized in Table \ref{datasets}. %\begin{flushleft}
 
\begin{table}[tbh]
\centering{}%
\begin{tabular}{cccc}
\hline 
{\small{}{}{}Dataset}  & {\small{}{}{}Market1501}  & {\small{}{}{}DukeMTMC}  & {\small{}{}{}CUHK03(D/L)}\tabularnewline
\hline 
{\small{}{}{}Identities}  & {\small{}{}{}1,501}  & {\small{}{}{}1,812}  & {\small{}{}{}1,360}\tabularnewline
{\small{}{}{}Bboxes}  & {\small{}{}{}32,668}  & {\small{}{}{}36,411}  & {\small{}{}{}13,164}\tabularnewline
{\small{}{}{}Camera}  & {\small{}{}{}6}  & {\small{}{}{}8}  & {\small{}{}{}6}\tabularnewline
{\small{}{}{}Train images}  & {\small{}{}{}12,936}  & {\small{}{}{}16,522}  & {\small{}{}{}7,365/7,368}\tabularnewline
{\small{}{}{}Train ids}  & {\small{}{}{}751}  & {\small{}{}{}702}  & {\small{}{}{}767}\tabularnewline
{\small{}{}{}Query images}  & {\small{}{}{}3,368}  & {\small{}{}{}2,228}  & {\small{}{}{}1,400}\tabularnewline
{\small{}{}{}Query ids}  & {\small{}{}{}750}  & {\small{}{}{}702}  & {\small{}{}{}700}\tabularnewline
{\small{}{}{}Gallery images}  & {\small{}{}{}19,732}  & {\small{}{}{}17,661}  & {\small{}{}{}5,332}\tabularnewline
\hline 
\end{tabular}\caption{ ReID Benchmark datasets used in our experiments.\label{datasets}}
\end{table}
%\par\end{flushleft}

We report Cumulative Matching Characteristics (CMC) at Rank-1 accuracy
and mean average precision (mAP) for all the three datasets. On Market-1501
dataset, we evaluate our method on two evaluation protocols:
single-query using one image as query, and multiple-query using more
than one images of one person in dataset to query. On CUHK03 dataset,
we follow the setting in \cite{Zhong2017Re} that splits
dataset into 767 training IDs and 700 test IDs, and report the performance on 
the test set with unknown IDs. Note that this protocol is different from the old 
setting  \cite{Mao2018Pyramid,Suh2018Part} with 1160 training IDs and 100 test
IDs, which the average of 20 trials of the random samples. Although the old setting 
often leads to  higher mAP or Rank-1 scores, it is not objective and suffers from the randomness. In this paper, we only compare with the results using the new setting as meaningful baselines. 
%some ReID methods \cite{Mao2018Pyramid,Suh2018Part} used the old
%protocol which splits the dataset into 1160 training IDs and 100 test
%IDs, and takes the average of 20 trials of the random samples. In
%contrast, PLA uses the new protocol \cite{Zhong2017Re} that splits
%dataset into 767 training IDs and 700 test IDs, which renders harder
%evaluation of ReID approaches. 

\subsection{Implementation Details}

Similar to \cite{Hermans2017In}, all images are resized to $256\times128$
resolution as a tradeoff between ReID efficiency and accuracy. For
online image augmentation, we first amplify by a factor 1.125, then
do a $256\times128$ random crop and a random horizontal flip in the
training process. Random erasing 
\cite{Zhong2017Random} is also used
which randomly ($p=0.5$) masks parts of person to simulate the cases
of occlusion and missing body parts in a dataset. We apply 10-crop
test time augmentation for evaluations. We use Principle Component
Analysis (PCA) to reduce the 2048-dim feature embedding to 512-dim
as post processing. As to mini-batch size, according to preliminary
experiments, increasing the number of images per person can benefit
the final accuracy of the trained model. Moreover, by using the proposed
progressive learning algorithm, images from the same person are required
to contain various levels of hard examples to optimize the training
process. Considering the GPU memory, we set $P=16$, $K=8$ to generate
image batches to feed into the model training process which is different
from \cite{Hermans2017In} with $P=18$ and $K=4$. We use Adam 
\cite{kingma2014adam}
as the optimizer and set $\beta_{1}=0.9$ within 150 epochs and $\beta_{1}=0.5$
for remaining epochs, and $\beta_{2}=0.999$. We adjust the learning
rate training schedule as proposed by \cite{Hermans2017In}: 
\begin{equation}
\alpha\text{\ensuremath{\left(e\right)}=}\begin{cases}
\alpha_{0} & \quad if\,e\leq e_{0}\\
\alpha_{0}\times0.001^{\frac{e-e_{0}}{e_{1}-e_{0}}} & \quad if\,e_{0}\leq e\leq e_{1}
\end{cases}
\end{equation}
where we set $\alpha_{0}=3\times10^{-4}$, $e_{0}=150$ epochs, and
$e_{1}=300$ epochs.

The simulation environments are given as follows: Ubuntu 16.04, Intel\textsuperscript{\textregistered}
Xeon\textsuperscript{\textregistered} CPU E5-2667 v4 @ 3.20GHz \texttimes{} 32,
64GB RAM, and NVIDIA\textsuperscript{\textregistered} GeForce\textsuperscript{\textregistered}
GTX 1080 Ti/PCIe/SSE2.

\subsection{Ablation Study }

\begin{figure}[tbh]
\centering\includegraphics[width=8cm, height=4cm]{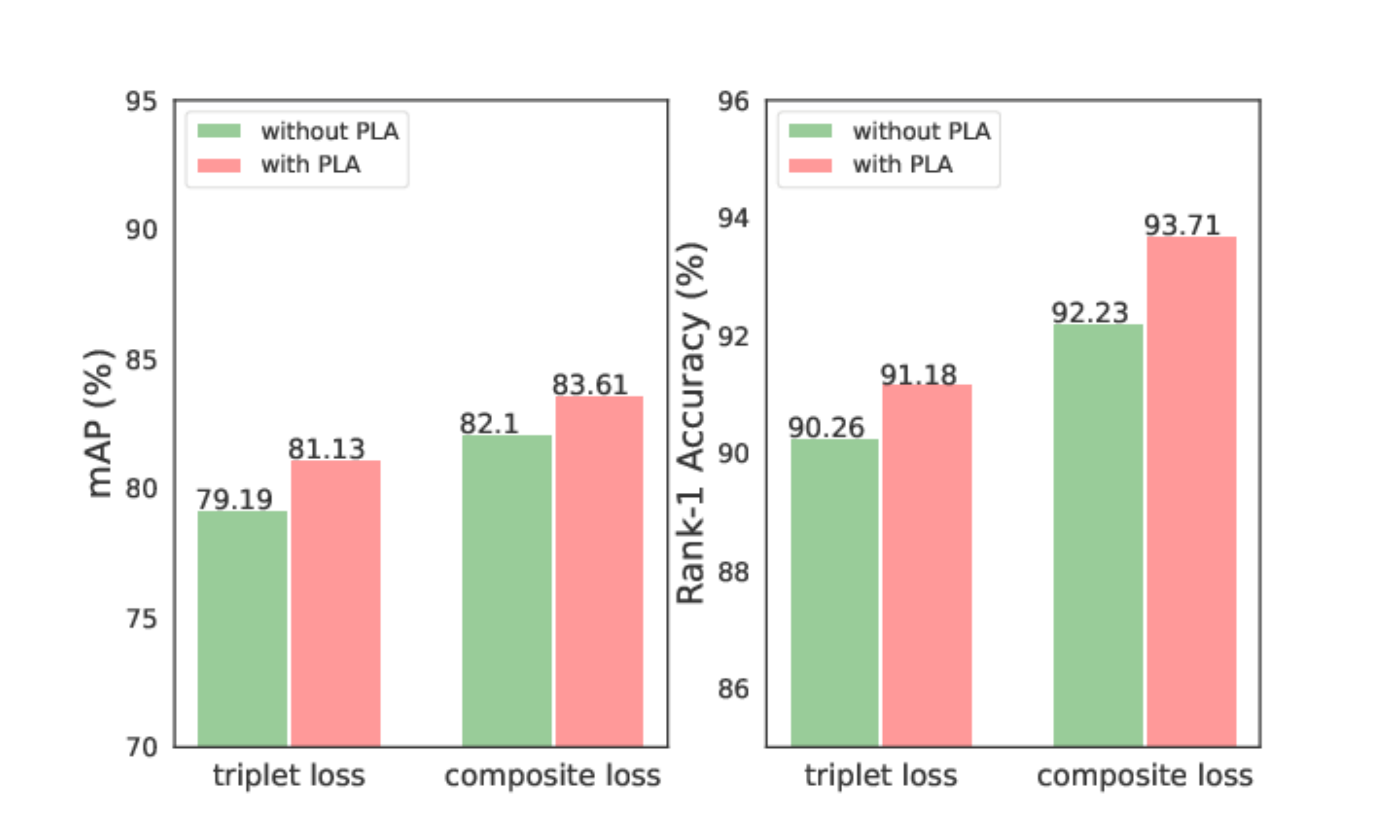} \caption{Using PLA on Market-1501 dataset. It can be seen that PLA consistently improves the performance of both triplet loss and  composite loss under every evaluation metrics.
\label{fig:ablation} }
\end{figure}

\begin{table*}[tb]
\centering{}%
\begin{tabular}{cccccccccccc}
\hline 
\multirow{2}{*}{{\small{}Category }} & \multirow{2}{*}{{\small{}Methods}} & \multicolumn{2}{c}{{\small{}Market1501(SQ)}} & \multicolumn{2}{c}{{\small{}Market1501(MQ)}} & \multicolumn{2}{c}{{\small{}CUHK03(D)}} & \multicolumn{2}{c}{{\small{}CUHK03(L)}} & \multicolumn{2}{c}{{\small{}DukeMTMC}}\tabularnewline
\cline{3-12} 
 &  & {\small{}mAP } & {\small{}Rank-1 } & {\small{}mAP } & {\small{}Rank-1 } & {\small{}mAP } & {\small{}Rank-1 } & {\small{}mAP } & {\small{}Rank-1 } & {\small{}mAP } & {\small{}Rank-1 }\tabularnewline
\multirow{6}{*}{{\small{}part}} & {\small{}HA-CNN\cite{Li2018Harmonious} } & {\small{}75.7} & {\small{}91.2} & {\small{}82.8 } & {\small{}93.8 } & {\small{}38.6 } & {\small{}41.7 } & {\small{}41.0 } & {\small{}44.4} & {\small{}63.8 } & {\small{}80.5}\tabularnewline
 & {\small{}Deep-Person\cite{Bai2017Deep} } & {\small{}79.6} & {\small{}92.3} & {\small{}85.1 } & {\small{}94.5 } & {\small{}- } & {\small{}- } & {\small{}- } & {\small{}-} & {\small{}64.8 } & {\small{}80.9}\tabularnewline
 & {\small{}PCB\cite{Sun2017Beyond} } & {\small{}77.4} & {\small{}92.3} & {\small{}- } & {\small{}- } & {\small{}54.2 } & {\small{}61.3 } & {\small{}- } & {\small{}-} & {\small{}66.1 } & {\small{}81.7}\tabularnewline
 & {\small{}PCB+RPP\cite{Sun2017Beyond} } & {\small{}81.6} & {\small{}93.8} & {\small{}- } & {\small{}- } & {\small{}57.5 } & {\small{}63.7 } & {\small{}- } & {\small{}-} & {\small{}69.2 } & {\small{}83.3}\tabularnewline
 & {\small{}Aligned-ReID\cite{zhang2017alignedreid} } & {\small{}82.3} & {\small{}92.6} & {\small{}- } & {\small{}- } & {\small{}- } & {\small{}- } & {\small{}- } & {\small{}-} & {\small{}- } & {\small{}- }\tabularnewline
 & {\small{}MGN\cite{Wang2018Learning} } & \textbf{\small{}86.9} & \textbf{\small{}95.7}{\small{} } & \textbf{\small{}90.7} & \textbf{\small{}96.9} & \textbf{\small{}66.0} & \textbf{\small{}66.8 } & \textbf{\small{}67.4 } & \textbf{\small{}68.0} & \textbf{\small{}78.4} & \textbf{\small{}88.7}\tabularnewline
\hline 
\multirow{5}{*}{{\small{}global}} & {\small{}SVDNet\cite{Sun2017SVDNet} } & {\small{}62.1} & {\small{}82.3} & {\small{}- } & {\small{}- } & {\small{}37.2 } & {\small{}41.5 } & {\small{}37.8 } & {\small{}40.9} & {\small{}56.8 } & {\small{}76.7}\tabularnewline
 & {\small{}TriNet\cite{Hermans2017In} } & {\small{}69.1} & {\small{}84.9} & {\small{}76.4 } & {\small{}90.5 } & {\small{}- } & {\small{}- } & {\small{}- } & {\small{}-} & {\small{}- } & {\small{}- }\tabularnewline
 & {\small{}GP-reid\cite{almazan2018re} } & {\small{}81.2} & {\small{}92.2} & {\small{}82.8 } & {\small{}93.8 } & {\small{}- } & {\small{}- } & {\small{}- } & {\small{}-} & \textbf{\small{}72.8} & \textbf{\small{}85.2}\tabularnewline
 & {\small{}DaRe\cite{wang2018resource} } & {\small{}74.2} & {\small{}88.5} & {\small{}- } & {\small{}- } & {\small{}58.1 } & {\small{}61.6 } & {\small{}60.2 } & {\small{}64.5} & {\small{}63.0 } & {\small{}79.1}\tabularnewline
 & \textcolor{black}{\small{}PLA} & \textbf{\small{}83.6} & \textbf{\small{}93.7} & \textbf{\small{}88.4} & \textbf{\small{}95.2} & \textbf{\small{}63.2} & \textbf{\small{}67.2} & \textbf{\small{}67.5} & \textbf{\small{}71.5} & {\small{}72.5 } & {\small{}84.3}\tabularnewline
\hline 
\multirow{4}{*}{{\small{}RK}} & {\small{}Trinet \cite{Hermans2017In} } & {\small{}81.1} & {\small{}86.7} & {\small{}87.2 } & {\small{}91.8 } & {\small{}- } & {\small{}- } & {\small{}- } & {\small{}- } & {\small{}- } & {\small{}- }\tabularnewline
 & {\small{}DaRe \cite{wang2018resource} } & {\small{}85.9} & {\small{}90.8} & {\small{}- } & {\small{}- } & {\small{}71.2 } & {\small{}69.8 } & {\small{}73.7 } & {\small{}72.9} & {\small{}79.6 } & {\small{}84.4}\tabularnewline
 & {\small{}MGN \cite{Wang2018Learning} } & \textbf{\small{}94.2} & \textbf{\small{}96.6} & \textbf{\small{}95.9} & \textbf{\small{}97.1} & {\small{}- } & {\small{}- } & {\small{}- } & {\small{}-} & {\small{}- } & {\small{}- }\tabularnewline
 & \textcolor{black}{\small{}PLA} & {\small{}89.4 } & {\small{}94.7} & {\small{}92.9} & {\small{}95.7} & \textbf{\small{}77.2} & \textbf{\small{}75.5} & \textbf{\small{}81.0} & \textbf{\small{}79.6} & \textbf{\small{}80.1} & \textbf{\small{}87.0}\tabularnewline
\hline 
\end{tabular}\caption{Comparing PLA with different global and part models on all datasets.
``RK'' stands for reranking. \label{state-of-the-art} }
\end{table*}

\textbf{Effects on Loss Concatenation.} In the proposed network structure, after global max pooling of the
ResNet-50 backbone, the network is split into two heads, one for triplet
loss and the other for cross-entropy loss. Considering that triplet
loss is a ranking function, while cross-entropy loss is a classification
target, their feature representations are different. Combining them
should have richer feature representation than each of them alone.
In our testing process, we used three variants of the network head:
2048-dim feature with cross-entropy loss only, 2048-dim feature with
triplet loss only, and the 2048-dim feature (concatenation of two
1024-dim feature embeddings). Fig.~\ref{fig:ablation} shows the effectiveness of loss concatenation on Market-1501 dataset in terms of mAP and Rank-1 accuracy. Coupling the feature concatenation method,
the proposed progressive learning algorithm brought significant improvements
$4.4\%$ for mAP and $3.5\%$ for Rank-1 compared to using triplet loss alone on Market-1501 dataset. 

\textbf{Effects on Progressive Learning.} Fig.~\ref{fig:ablation} shows the effectiveness of progressive learning on Market-1501 dataset. With the proposed PLA, the triplet feature
embedding achieved mAP and Rank-1 accuracy of $81.1\%$ and $91.2\%$,
respectively, outperforming the accuracies using each type of loss
with 2048-dim feature without PLA. We
found progressive learning and concatenated feature embedding, when
used together, produced the best results. This indicates the progressive
learning is effective to learn more accurate and robust representation
of person identities.

\begin{figure}[tbh]
    \centering\includegraphics[scale=0.28]{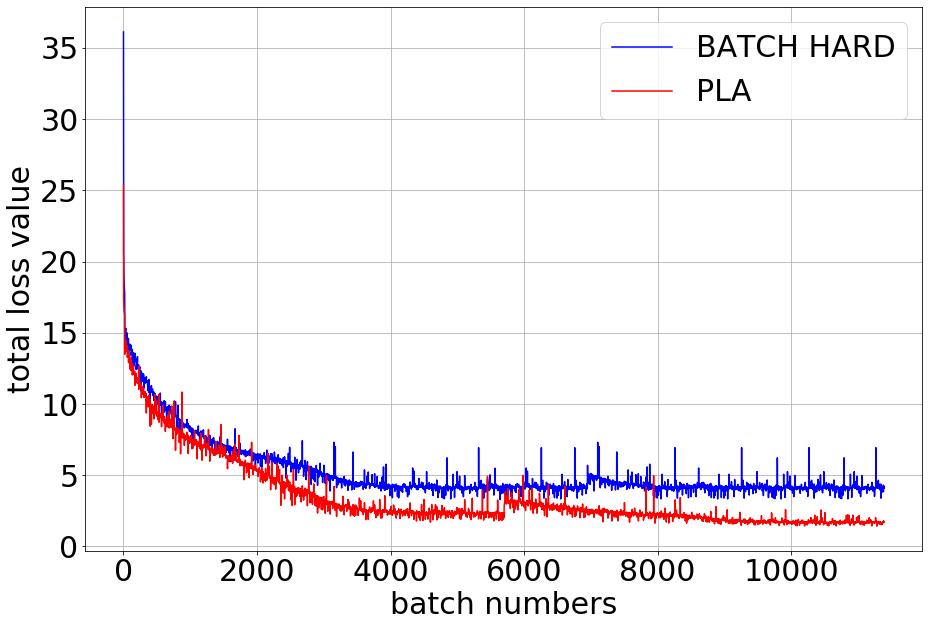}
 \caption{The blue line is for batch hard training loss curve; The red line is for PLA training loss curve. \label{fig:loss_comparison}}
\end{figure}

\textbf{Effectiveness on Training Convergence.} To illustrate the
robustness of our loss function on the convergence performance, we use the same ResNet-50 backbone as the network architecture and train it on Market1501 using two methods: 1) the proposed PLA training method; 2) batch hard training \cite{Hermans2017In}. As shown in
Fig. \ref{fig:loss_comparison}, the red curve shows loss values of our method, which entails two stages compared with batch hard training method shown as the blue curve. It is observed that in the first training stage (first 5,700 batches), our loss converges faster than batch hard loss as the proposed method trains easy and medium samples at the beginning. When hard samples are introduced to the proposed method, it comes to the second stage. The loss is significantly reduced further using PLA, leading to the final improvement, while the loss of hard samples cannot be reduced any more. Note that in the beginning of the second training stage, the red curve has a sudden jump because harder samples are introduced; however, the loss value gradually dropped until achieving the lowest ever one. The
results illustrate that learning progressively by the level of hardness is beneficial to model convergence and achieves lower loss.

\subsection{Compare with the State-of-the-arts}

We compare the proposed ReID approach with various ReID approaches
based on both global models and part models. The global model based
approaches include: SVDNet \cite{Sun2017SVDNet}, TriNet \cite{Hermans2017In},
\textcolor{black}{GP-reid \cite{almazan2018re}}, and ResNet-50 based
DaRe \cite{wang2018resource}. The part model based approaches include:
HA-CNN \cite{Li2018Harmonious}, PCB \cite{Sun2017Beyond}, Aligned-ReID
\cite{zhang2017alignedreid}, Deep-Person \cite{Bai2017Deep} and
MGN \cite{Wang2018Learning}.

The comparison results over state-of-the-art methods with global models
are shown in Table \ref{state-of-the-art}. Without the reranking
technique, GP-reid \cite{almazan2018re} achieved a good performance
as it properly makes use of many existing techniques. The proposed
PLA consistently outperforms all other global feature models by a
large margin without reranking technique, except for GP-reid. PLA
outperforms GP-reid in terms of mAP accuracy by $2.4\%$ and $5.6\%$
on Market-1501 dataset with single query mode and multiple query mode,
respectively. However, GP-reid slightly outperforms PLA in terms of
mAP and Rank-1 accuracy by $0.3\%$ and $0.9\%$, respectively, on
DukeMTMC dataset. Note that in this work the learning rate schedule\textcolor{black}{{}
is uniform in all experiments and not specially tailored for DukeMTMC
dataset. We can observe that the proposed PLA consistently outperforms
all other methods with reranking by a large margin. PLA outperforms
DaRe in terms of mAP by $9.4\%$, $5.1\%$, $7.3\%$ and $9.5\%$
without reranking on Market-1501(SQ), CUHK03(D), CUHK03(L) and DukeMTMC
datasets respectively. Overall, the proposed PLA has the best performance
over other global models.}

Although the proposed approach is within the scope of global feature
model, we also compare it with the accuracies produced by multi-branch
part models, as shown in Table \ref{state-of-the-art} for all datasets.
In Table \ref{state-of-the-art}, it is obvious that MGN achieves
the highest accuracies among all part models on Market-1501 dataset
without reranking technique. It is evident that the proposed PLA outperforms
all previous works except for MGN. In terms of Rank-1 accuracy performance
without reranking, PLA is overall comparable with MGN in terms of
Rank-1 accuracy: MGN exceeds PLA by $2\%$ and $4.3\%$ on Market-1501(SQ)
and DukeMTMC dataset respectively, while MGN is inferior than PLA
by $0.4\%$ and $3.5\%$ on CUHK03(D) and CUHK03(L) datasets. We can
also observe that with reranking, PLA outperforms DaRe by a large
margin achieving the highest mAP and Rank-1 values ever publicly reported
on the CUHK03(D) and CUHK03(L) datasets. %However, MGN divides the network into multiple branches before global pooling, thus introducing a lot more CNN weights to improve the accuracy at the cost of high computation and memory resources.

\subsection{Computational Cost and Inference Memory}

The proposed PLA is designed for resource-aware ReID, and a good trade
off between accuracy and efficiency is expected. The computational
cost is reported with respect to the total number of basic operations,
i.e., multiplications and additions (Mul-Add), which is proportional
to the actual running timing. Reranking is not included in the Mul-Add
calculation. HA-CNN \cite{Li2018Harmonious}, ResNet-50 based DaRe
\cite{wang2018resource}, PCB(RPP) \cite{Sun2017Beyond} and MGN \cite{Wang2018Learning}
are chosen to be compared with the proposed PLA for their good performance.

\begin{figure}[tbh]
  \begin{subfigure}{4cm}
    \centering\includegraphics[scale=0.28]{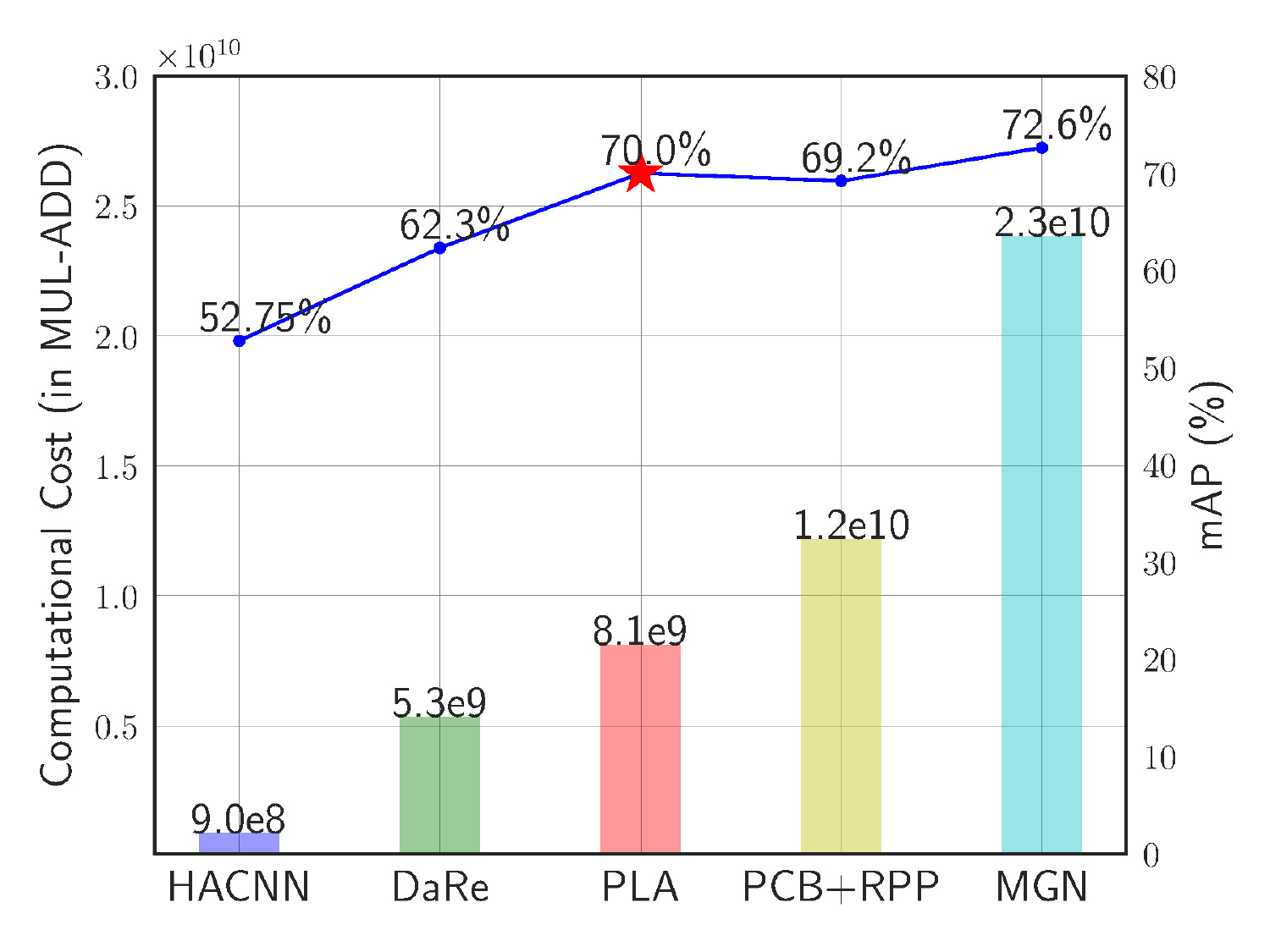}
    \caption{}
  \end{subfigure}
  \begin{subfigure}{4cm}
    \centering\includegraphics[scale=0.28]{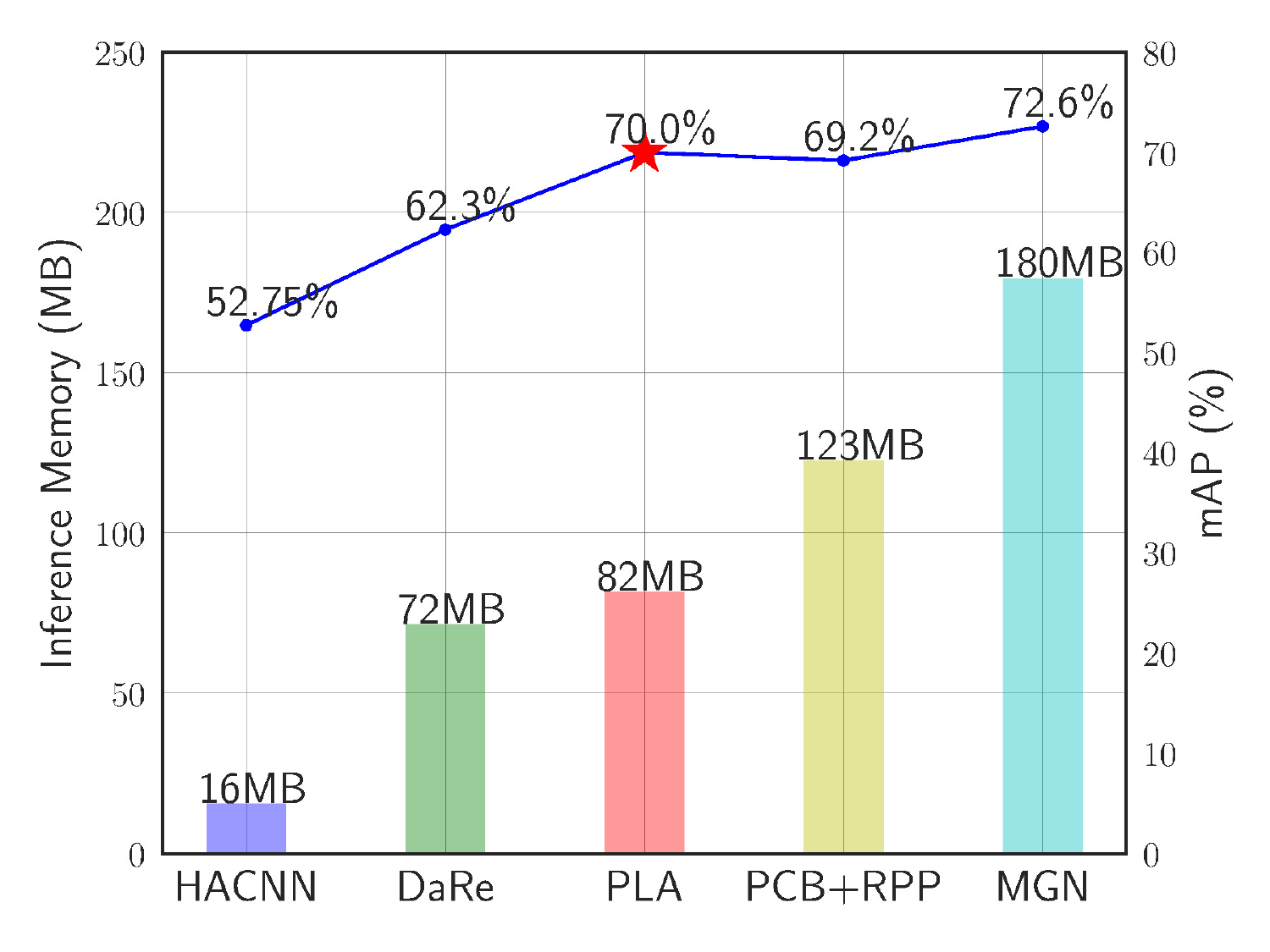}
    %\caption{Accuracy vs. inference memory (MB)}
    \caption{}
  \end{subfigure}
 \caption{(a) Accuracy vs. computation cost (number of Mul-Add); (b) Accuracy vs. inference memory (MB). Accuracy is reported
as the average mAP on all datasets. \label{fig:resources}}
\end{figure}

The graph of accuracy vs. efficiency is shown in Fig. \ref{fig:resources} (a).
Accuracy reported is the average mAP on Market-1501, CUHK03(D), CUHK03(L)
and DukeMTMC datasets without reranking. We observe that the proposed
PLA achieves significantly higher accuracies than HA-CNN and DaRe.
In particular, PLA achieves slightly higher accuracy than PCB which
is a strong part model baseline for ReID, yet saving more than $30\%$
computation cost. MGN outperforms PLA and PCB in accuracy by a small
margin, however, it has a tremendous Mul-Add of roughly $2.3\times10^{10}$
which is about 3 times of PLA.

% \begin{figure}[tbh]
% \centering\includegraphics[scale=0.35]{plot_bar_cost.pdf} \caption{Accuracy vs. computation cost (number of Mul-Add). Accuracy is reported
% as the average mAP on all datasets. \label{fig:resource-computation} }
% \end{figure}
The graph of accuracy vs. inference memory is shown in Fig. \ref{fig:resources} (b).
It is evident that the trade off situation is similar to that in Fig.
\ref{fig:resources} (a). Although HA-CNN is a multi-branch
part model, the input size is fairly small compared with other approaches,
making it the most lightweight approach. HA-CNN shows good accuracy
on Market-1501 dataset but exhibits rather low accuracies on CUHK
and DukeMTMC datasets, as shown in Table \ref{state-of-the-art}.
MGN outperforms all others in terms of accuracy, but it is more than
2 times of the PLA inference memory. Accordingly, it requires at least
11.6GB memory for inference with a commonly used batch size of 64,
which even cannot be loaded using GTX 1080Ti (11.17GB). Note that
even though network acceleration techniques  TensorRT \cite{TensorRT}
and network compression techniques
%\eg \cite{wang2018exploring}
can be used to save Mul-Add and reduce inference memory, respectively,
for strong part models including PCB and MGN, they will also benefit
PLA by similar acceleration and compression ratios since ResNet-50
is the backbone of all these networks.

It is evident that in Fig. \ref{fig:resources},
our method is outstanding in its accuracy and resource balance: the
models with less resource requirements exhibit obvious lower or unstable
accuracy while those with slightly higher accuracy will consume unnecessarily
much more computational cost and inference memory.

\section{Conclusion}
Different from the recent trend of employing part models to improve
the accuracy at the cost of more computation and memory, this paper
developed a novel optimization algorithm to find efficient models
in a single main branch network with competitive performance. The
core idea is to design a progressive triplet loss to explore the power
of deep CNN representation, which is efficient for model inference.
Compared to the most state-of-the-art algorithms, e.g., MGN, which
produces slightly higher accuracy than PLA on average, PLA saves 65\%
computational cost and 54\% inference memory. Experimental
results show that the proposed PLA achieves the excellent tradeoff between
accuracy and efficiency, and we hope this work
could motive more research work to find efficient models with high
performance and lower consumption.

\section{Acknowledgments}
The authors would like to thank Liangliang Cao for his constructive feedback, and the anonymous reviewers for their insightful and valuable comments.


%\section*{Acknowledgment}




%\bibliographystyle{IEEEtran}
%\begin{thebibliography}{1}


%\end{thebibliography}
\bibliographystyle{IEEEtran}
\bibliography{reid}

\end{document}